\documentclass[letterpaper]{article} % DO NOT CHANGE THIS
\usepackage{aaai2026}  % DO NOT CHANGE THIS
\usepackage{times}  % DO NOT CHANGE THIS
\usepackage{helvet}  % DO NOT CHANGE THIS
\usepackage{courier}  % DO NOT CHANGE THIS
\usepackage[hyphens]{url}  % DO NOT CHANGE THIS
\usepackage{graphicx} % DO NOT CHANGE THIS
\usepackage{natbib}  % DO NOT CHANGE THIS AND DO NOT ADD ANY OPTIONS TO IT
\usepackage{caption} % DO NOT CHANGE THIS AND DO NOT ADD ANY OPTIONS TO IT
\usepackage{algorithm}
\usepackage{algorithmic}
\usepackage{amsmath}
\usepackage{amssymb}

\usepackage{newfloat}
\usepackage{listings}
\DeclareCaptionStyle{ruled}{labelfont=normalfont,labelsep=colon,strut=off} % DO NOT CHANGE THIS
\floatstyle{ruled}
\newfloat{listing}{tb}{lst}{}
\floatname{listing}{Listing}
\title{Making Every Head Count: Sparse Attention \\Without the Speed-Performance Trade-off}
\author{
    Mingkuan Zhao\textsuperscript{\rm 1}\equalcontrib,
    Wentao Hu\textsuperscript{\rm 1}\equalcontrib,
    Jiayin Wang\textsuperscript{\rm 1},
    Xin Lai\textsuperscript{\rm 1},
    Tianchen Huang\textsuperscript{\rm 2},\\
    Yuheng Min\textsuperscript{\rm 3},
    Rui Yan\textsuperscript{\rm 4},
    Xiaoyan Zhu\textsuperscript{\rm 1}\thanks{Corresponding author.}
}
\affiliations{
    \textsuperscript{\rm 1}Xi'an Jiaotong University\\
    \textsuperscript{\rm 2}University of Science and Technology of China\\
    \textsuperscript{\rm 3}Tsinghua University\\
    \textsuperscript{\rm 4}University of California, San Diego\\
    mingkuanzhao@stu.xjtu.edu.cn, wentao\_hu@stu.xjtu.edu.cn, wangjiayin@mail.xjtu.edu.cn, laixin@xjtu.edu.cn, \\
    tchuang@mail.ustc.edu.cn, minyh24@mails.tsinghua.edu.cn, jerryyan24@outlook.com, zhu.xy@xjtu.edu.cn
}

\usepackage{bibentry}
\begin{document}

\maketitle

\begin{abstract}
The design of Large Language Models (LLMs) has long been hampered by a fundamental conflict within their core attention mechanism: its remarkable expressivity is built upon a computational complexity of O(H·N²) that grows quadratically with the context size (N) and linearly with the number of heads (H). This standard implementation harbors significant computational redundancy, as all heads independently compute attention over the same sequence space. Existing sparse methods, meanwhile, often trade information integrity for computational efficiency. To resolve this efficiency-performance trade-off, we propose SPAttention, whose core contribution is the introduction of a new paradigm we term Principled Structural Sparsity. SPAttention does not merely drop connections but instead reorganizes the computational task by partitioning the total attention workload into balanced, non-overlapping distance bands, assigning each head a unique segment. This approach transforms the multi-head attention mechanism from H independent O(N²) computations into a single, collaborative O(N²) computation, fundamentally reducing complexity by a factor of H. The structured inductive bias compels functional specialization among heads, enabling a more efficient allocation of computational resources from redundant modeling to distinct dependencies across the entire sequence span. Extensive empirical validation on the OLMoE-1B-7B and 0.25B-1.75B model series demonstrates that while delivering an approximately two-fold increase in training throughput, its performance is on par with standard dense attention, even surpassing it on select key metrics, while consistently outperforming representative sparse attention methods including Longformer, Reformer, and BigBird across all evaluation metrics. Our work demonstrates that thoughtfully designed structural sparsity can serve as an effective inductive bias that simultaneously improves both computational efficiency and model performance, opening a new avenue for the architectural design of next-generation, high-performance LLMs.
\end{abstract}

\begin{links}
    \link{Code}{https://github.com/Harry-Miral/MEHC}
    %\link{Datasets}{https ://huggingface.co/datasets/mlfoundations/dclm-baseline-1.0}
    %\link{Extended version}{https://aaai.org/example/extended-version}
\end{links}

\section{Introduction}

The advancement of Large Language Models (LLMs) is fundamentally constrained by their core component, the self-attention mechanism. The severity of this bottleneck, a core challenge for Transformer architectures across modalities \cite{dosovitskiy2020image}, lies not only in its quadratic computational growth with sequence length ($N$) but, more profoundly, in the computational redundancy inherent to standard Multi-Head Attention (MHA). With a complexity of $O(H \cdot N^2)$, the standard model compels each of its $H$ heads to independently perform a full $O(N^2)$ attention computation across the entire context. This design is structurally inefficient, as extensive research has revealed that many heads in trained models spontaneously converge to functionally similar and simple patterns, such as merely attending to adjacent tokens \cite{clark2019doesbertlookat, voita2019analyzingmultiheadselfattentionspecialized}. This highlights a critical inefficiency: a substantial portion of the $H$-fold computational budget is spent on learning repetitive, basic functions, rather than being fully leveraged for modeling a diverse spectrum of dependencies.

Against this backdrop, existing optimization paths have failed to provide a fundamental solution. At the algorithmic level, sparse attention methods \cite{child2019generatinglongsequencessparse, beltagy2020longformerlongdocumenttransformer, kitaev2020reformerefficienttransformer, NEURIPS2020_c8512d14} often reduce computation via hard pruning, but this frequently sacrifices the model's intrinsic expressivity and risks critical information loss. Head pruning methods \cite{voita2019analyzingmultiheadselfattentionspecialized, michel2019sixteen} are post-hoc remedies that address redundancy only after training is complete. Such approaches incur the full computational cost during the training phase and fail to resolve the inefficient resource allocation at its source. System-level optimizations like FlashAttention \cite{NEURIPS2022_67d57c32}, while achieving substantial speedups through hardware-aware implementation, are orthogonal and complementary to our work; they optimize the execution of the $O(H \cdot N^2)$ computation but do not alter its fundamental asymptotic complexity or structural inefficiency.

This landscape motivates a paradigm shift from post-hoc remedies to a priori architectural design. In response to this challenge, we propose SPAttention (Sparse Patterned Attention), a novel attention mechanism based on what we term Principled Structural Sparsity. The design philosophy of SPAttention is not to simply 'drop' connections but to 'reorganize and specialize' the total computational workload. By partitioning the entire causal attention distance spectrum into $H$ balanced, non-overlapping bands, SPAttention assigns each head a unique and exclusive area of responsibility. This introduces an effective structured inductive bias that compels functional specialization, enabling a more efficient allocation of the model's finite capacity. Computational resources are liberated from redundant local modeling and precisely redirected to capturing distinct dependencies across the entire sequence span. This design elegantly reduces the computational complexity from $O(H \cdot N^2)$ to $O(N^2)$ while maintaining a complete information pathway.

To validate our design, we integrated SPAttention into the OLMoE architecture \cite{muennighoff2025olmoeopenmixtureofexpertslanguage}. Our experiments, conducted on models up to the 7B scale, demonstrate that SPAttention delivers an approximately two-fold increase in measured training throughput. Remarkably, this efficiency gain does not come at the cost of performance; SPAttention's performance matches that of standard dense attention and even surpasses it on select key metrics. Our work demonstrates that thoughtfully designed structural sparsity is not a performance compromise but a powerful inductive bias that can simultaneously unlock both superior computational efficiency and model performance, opening a new avenue for the architectural design of next-generation LLMs, particularly in tackling the challenge of expensive pre-training.

\section{Related Work}

To overcome the quadratic complexity bottleneck of self-attention, research has proceeded along several axes, including algorithmic sparsity, head specialization, and system-level optimization. SPAttention contributes a novel approach to algorithmic sparsity that enforces head specialization by design.

A prominent line of work in algorithmic sparsity involves fixed or hybrid attention patterns \cite{beltagy2020longformerlongdocumenttransformer, NEURIPS2020_c8512d14, ainslie2020etc, child2019generating, tay2020long, liu2023blockwise}. Within this paradigm, representative methods like Longformer \cite{beltagy2020longformerlongdocumenttransformer}, BigBird \cite{NEURIPS2020_c8512d14}, and ETC \cite{ainslie2020etc} employ a hybrid of local windowed attention and a few global attention tokens. While this reduces computation, it creates an information bottleneck by forcing all non-local dependencies through a few global tokens, which places a heavy representational burden on them and risks critical information loss.

More closely related to our work are methods that differentiate computational roles across heads or learn data-dependent sparsity patterns \cite{child2019generating, kitaev2020reformerefficienttransformer, roy2020efficientcontentbasedsparseattention, correia2019adaptively, tay2020sparse}. The Sparse Transformer \cite{child2019generating} was an early attempt, proposing a mix of fixed patterns where some heads attend locally and others use strided attention. However, this still permits redundancy within head groups sharing the same pattern (e.g., all local heads) and the strided patterns create information "gaps" at the single-layer level. Other methods pursue adaptive sparsity; for instance, Reformer \cite{kitaev2020reformerefficienttransformer} uses LSH-based attention, and others like Routing Transformer \cite{roy2020efficientcontentbasedsparseattention} learn data-dependent connections. However, their irregular memory access patterns are difficult to optimize on parallel hardware, often failing to translate theoretical FLOPs reduction into practical throughput gains. In sharp contrast, SPAttention assigns a \textit{unique}, \textit{contiguous}, and \textit{non-overlapping} distance band to each head, creating a seamless partition of the dependency spectrum. This ensures complete information coverage while enforcing functional specialization. Furthermore, its highly regular block-sparse structure is exceptionally hardware-friendly.

Another paradigm seeks to guide head specialization without enforcing sparsity, often by incorporating various forms of positional information as a soft inductive bias \cite{vaswani2017attention, raffel2020exploring, press2021train, su2024roformer}. Prominent examples include T5's relative position bias \cite{raffel2020exploring} and ALiBi \cite{press2021train}, which add a distance-dependent penalty to the attention logits. Crucially, these "soft" constraints do not reduce the fundamental $O(H \cdot N^2)$ computation, as they still require calculating the full attention matrix. Taking a different approach, Talking-Heads Attention \cite{shazeer2020talkingheadsattention} facilitates inter-head communication by adding extra transformations, but does so at the cost of increased computational complexity and parameters. SPAttention's "hard" structural constraint is fundamentally different. It is the very mechanism that simultaneously guarantees head diversity and enables the reduction of computational complexity to $O(N^2)$, offering both a powerful inductive bias and a massive efficiency gain.

Finally, our work is informed by analyses of Transformer internals and complemented by system-level optimizations. A body of research has provided post-hoc diagnoses of MHA's functional redundancy, revealing that many heads are functionally similar or can be pruned without significant performance loss \cite{clark2019doesbertlookat, voita2019analyzingmultiheadselfattentionspecialized, michel2019sixteen}. Orthogonally, system-level optimizations that are complementary to our work, such as FlashAttention \cite{NEURIPS2022_67d57c32} for training and others for inference \cite{pope2023efficiently}, optimize the execution efficiency on hardware but do not alter the fundamental algorithm. In fact, SPAttention's regular, block-sparse pattern is exceptionally well-suited for such IO-aware kernels.

\section{Methodology}

\begin{figure*}[t]
	\centering
	\includegraphics[width=0.8\textwidth]{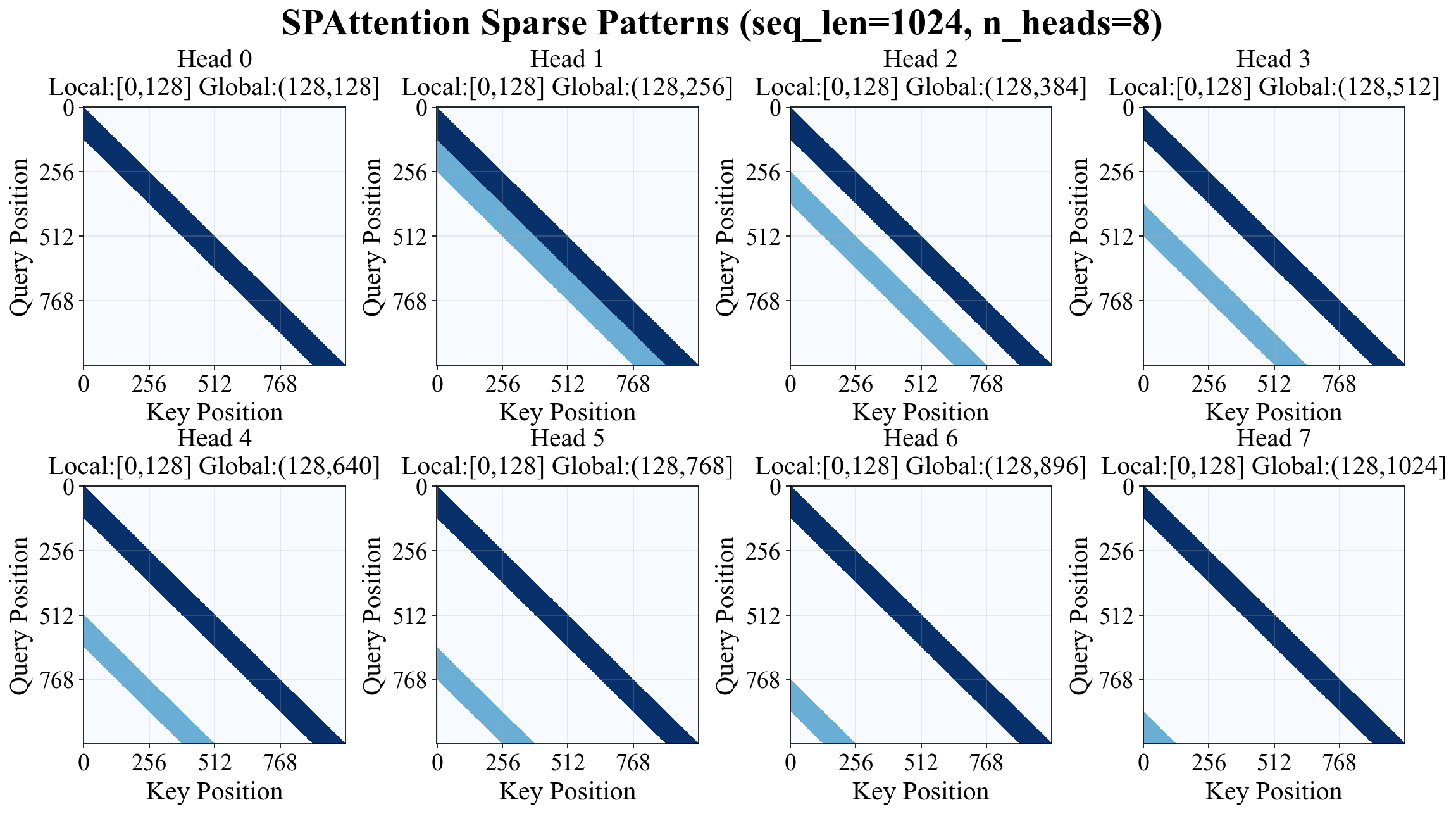}
	\caption{An illustration of the SPAttention sparse patterns for a sequence of length $N=1024$ with $H=8$ heads. Each subplot shows the attention pattern for an individual head. The entire causal attention distance spectrum is partitioned into eight contiguous, non-overlapping bands through \textbf{Balanced Distance Partitioning}, with each head assigned to exactly one band of width $\lfloor N/H \rfloor$ or $\lceil N/H \rceil$. This design guarantees complete, gapless information coverage while compelling different heads to specialize on distinct distance ranges—from immediate neighbors (Head 0) to long-range dependencies (Head 7).}
	\label{fig:spattention_pattern}
\end{figure*}

This section details the core mechanism of SPAttention. We first introduce its design philosophy and conceptual framework, then provide its rigorous formal definition, and finally, offer a multi-faceted theoretical analysis, supported by empirical validation, to explain why this structured sparse design achieves dual improvements in computational efficiency and model performance.

\subsection{Conceptual Framework: From Redundancy to Specialization}

The fundamental flaw of standard Multi-Head Attention (MHA) lies in its structural and computational redundancy. With a complexity of $O(H \cdot N^2)$, it forces each of its $H$ heads to independently solve the same task, leading to functionally overlapping patterns and wasted computation. The design philosophy of SPAttention is to address this issue at its root. Instead of merely pruning connections, we introduce a powerful, domain-agnostic structured inductive bias to ``reorganize and specialize'' the entire computational workload. We call this paradigm Principled Structural Sparsity, implemented through Balanced Distance Partitioning.

This principle is not arbitrary but founded on three core tenets: completeness, exclusivity, and balance. As illustrated in Figure~\ref{fig:spattention_pattern}, our strategy elegantly divides the entire causal attention distance spectrum into $H$ bands.
\textbf{1. Completeness:} The bands are contiguous and their union covers all possible causal distances from $0$ to $N-1$. This ensures that, unlike methods with strided or random gaps, a complete and unbroken information pathway is always maintained.
\textbf{2. Exclusivity:} The bands are mutually exclusive and non-overlapping. Each head is assigned exactly one band, making it a designated specialist for a specific range of dependencies. This hard constraint eradicates functional redundancy by design. For instance, the head assigned the most local band is forced to master short-range syntactical relationships, while a head assigned a distant band can specialize in capturing long-range thematic coherence, a clear division of labor.
\textbf{3. Balance:} The partitioning algorithm ensures each head receives a nearly identical workload (band widths differ by at most one token). This perfect load-balancing is critical for maximizing parallelism on modern hardware.

Through this design, SPAttention transforms MHA from a redundant committee of generalists into a highly efficient, synergistic assembly of specialists. It eliminates the computational waste inherent in standard MHA and converts the inefficient, passively-emerging pattern overlap into an active and deliberate \textbf{Functional Specialization}.

\textbf{Hyperparameter-Free Design.} A distinctive advantage of SPAttention is its elimination of manual hyperparameter tuning. Unlike existing sparse methods that require careful adjustment of window sizes, stride lengths, or global token ratios, SPAttention's balanced partitioning is deterministically defined by the model's architectural parameters: sequence length $N$ and number of heads $H$. This hyperparameter-free nature simplifies implementation, removes task-specific tuning overhead, and ensures robust, consistent performance across diverse applications.

\subsection{Formal Definition}

The output of the $h$-th head in a standard MHA \cite{vaswani2017attention} is given by:
\begin{equation}
	\text{Attention}_h(Q_h, K_h, V_h) = \text{softmax}\left(\frac{Q_h K_h^T}{\sqrt{d_k}} + M_h\right)V_h
\end{equation}
where the mask $M_h$ determines the connectivity. The core innovation of SPAttention lies in the construction of a set of masks $\{M_h\}_{h=0}^{H-1}$ that partitions the computational workload.

For a sequence of length $N$ and $H$ attention heads, we divide the causal attention distance spectrum $\{0, 1, \ldots, N-1\}$ into $H$ contiguous and non-overlapping bands. Let the base partition width be $W_{\text{base}} = \lfloor N / H \rfloor$ and the remainder be $R = N \bmod H$. To ensure a perfectly balanced load distribution, the first $R$ heads are assigned a slightly wider band:
\begin{align}
	W_h &= W_{\text{base}} + \mathbb{I}(h < R) \label{eq:width}\\
	S_h &= h \cdot W_{\text{base}} + \min(h, R) \label{eq:start}
\end{align}
where $W_h$ is the width and $S_h$ is the starting distance of the band assigned to head $h \in \{0, 1, \ldots, H-1\}$, and $\mathbb{I}(\cdot)$ is the indicator function.

The attention mask $M_h$ for each head is then constructed based on this partitioning. A connection from query position $i$ to key position $j$ is allowed for head $h$ if and only if their distance $i-j$ falls within the head's designated band:
\begin{equation}
	\text{Allow}_h(i, j) \equiv (j \leq i) \land (S_h \leq i-j < S_h + W_h) \label{eq:allow}
\end{equation}
The corresponding mask entry $M_h(i, j)$ is $0$ if this condition is met, and $-\infty$ otherwise. A crucial property of this design is that the union of allowed connections across all heads covers all causal token pairs, ensuring a complete information pathway with no gaps:
\begin{equation}
	\bigcup_{h=0}^{H-1} \{(i,j) \mid \text{Allow}_h(i, j)\} = \{(i,j) \mid j \leq i\}
\end{equation}

\subsection{Theoretical Analysis}

The effectiveness of SPAttention is rooted in solid theoretical foundations, which we analyze from the perspectives of computational complexity and the power of its inductive bias.

\subsubsection{Efficiency through Complexity Reduction and Hardware Co-design.}

The primary source of SPAttention's efficiency gain is a fundamental reduction in computational complexity. Standard Multi-Head Attention involves $H$ independent attention computations, each operating over the full $O(N^2)$ space of causal connections. This results in a total complexity of:
\begin{equation}
	\text{Complexity}_{\text{Standard}} = H \cdot O(N^2) = O(H \cdot N^2)
\end{equation}

SPAttention redesigns this process. Instead of $H$ redundant computations, it performs a single, partitioned computation distributed across all heads. The total computational complexity is the sum of the work done by each specialized head:
\begin{align}
	\text{Complexity}_{\text{SPA}} &= \sum_{h=0}^{H-1} O(N \cdot W_h) = O\left(N \cdot \sum_{h=0}^{H-1} W_h\right) \notag \\
	&= O(N \cdot N) = O(N^2)
\end{align}
This constitutes a theoretical reduction in floating-point operations (FLOPs) by a factor of $H$.

While the theoretical speedup is $H$, the measured training throughput improvement is approximately 2$\times$. This is explained by three key factors. First, according to Amdahl's Law, the total speedup is limited by the proportion of time spent on the attention calculation itself, as other components like the Feed-Forward Networks (FFNs) remain unchanged. Second, the implementation of any sparse attention, including ours, incurs a degree of overhead for mask management and block reorganization that is not present in the dense counterpart. Finally, and most critically, SPAttention's design represents an ideal instance of algorithm-hardware co-design. Its highly regular, contiguous block-sparse pattern is exceptionally friendly to modern hardware and IO-aware kernels like FlashAttention. This crucial property allows the theoretical FLOPs reduction to be translated into substantial real-world speedup despite the aforementioned limitations—a feat that eludes sparse methods with irregular memory access patterns.

\subsubsection{Performance via Inductive Bias and Regularization.}

SPAttention's strong performance stems from the powerful inductive bias it imposes, which acts as a form of structural regularization. By design, the attention supports of any two distinct heads are disjoint:
\begin{equation}
	\mathcal{J}_{i,h} \cap \mathcal{J}_{i,h'} = \emptyset, \quad \forall h \neq h', \forall i
\end{equation}
where $\mathcal{J}_{i,h} = \{j \mid \text{Allow}_h(i, j)\}$ is the set of allowed keys for query $i$ in head $h$. This hard structural constraint fundamentally prevents heads from converging to similar functions, forcing them to specialize. This structure also imposes a natural entropy regularization on each head, as the maximum Shannon entropy is bounded by $H_{\max}(p_h) \approx \log(N/H)$, significantly smaller than the $\log(i+1)$ bound for standard attention.

To empirically validate these theoretical predictions, we conducted controlled simulation experiments using 4-layer, 8-head Transformer models (see Appendix A for methodology). The results provide strong quantitative backing for our analysis. The enforced structural constraints dramatically increase head specialization: standard attention shows near-zero head diversity ($\sigma \approx 0.0000$--$0.0005$), empirically confirming the functional redundancy problem, while SPAttention demonstrates consistent high diversity ($\sigma \approx 0.1845$--$0.1847$), representing over a 300$\times$ enhancement. This validates that our disjoint support design successfully prevents functional convergence. Furthermore, the measured entropy values align with our theoretical bounds: standard attention achieves an average entropy of $\bar{H} = 4.5461$, whereas SPAttention's is $\bar{H} = 3.6276$, a 20.20\% reduction that confirms the effect of its implicit regularization. These findings establish a direct empirical link between SPAttention's structure and its desirable emergent properties.

\begin{table*}[t]
	\centering
	\begin{tabular}{lccccccc}
		\hline
		\textbf{Method} & \textbf{Winogrande} & \textbf{COPA} & \textbf{STEM} & \textbf{SocSci} & \textbf{STEM (5-shot)} & \textbf{Humanities (5-shot)} & \textbf{Average} \\ \hline
		\multicolumn{8}{c}{\textbf{OLMoE 0.25B-1.75B}} \\ \hline
		Standard Attention & 0.5130 & 0.6400 & 0.2528 & \textbf{0.2640} & 0.2440 & \textbf{0.2520} & 0.3610 \\
		SPAttention   & \textbf{0.5154} & \textbf{0.6600} & \textbf{0.2666} & 0.2552 & \textbf{0.2664} & 0.2468 & \textbf{0.3684} \\ \hline
		\multicolumn{8}{c}{\textbf{OLMoE 1B-7B}} \\ \hline
		Standard Attention & \textbf{0.5040} & 0.5900 & 0.2926 & 0.2181 & 0.1809 & 0.2553 & 0.3402 \\
		SPAttention   & 0.4956 & \textbf{0.6100} & \textbf{0.3022} & \textbf{0.2533} & \textbf{0.2000} & \textbf{0.2578} & \textbf{0.3532} \\ \hline
	\end{tabular}
	\caption{Performance Comparison across different model scales. The MMLU sub-tasks are evaluated under zero-shot (e.g., STEM) and five-shot settings. Best results within each scale are highlighted in bold.}
	\label{tab:main_results}
\end{table*}

\section{Experiments}

To comprehensively evaluate the effectiveness of SPAttention, we conducted a series of extensive experiments. This chapter first details our experimental setup, then presents and analyzes our main results across two model scales, practical throughput tests, and in-depth ablation studies. Our findings demonstrate that SPAttention achieves significant efficiency gains and performance improvements without trade-offs.

\subsection{Experimental Setup}

The core of all our experiments is large-scale pre-training based on the OLMoE framework \cite{muennighoff2025olmoeopenmixtureofexpertslanguage}. We recognize that pre-training large language models is a computationally prohibitive endeavor, demanding immense GPU resources and time.
Therefore, our contribution is not merely proposing a novel attention mechanism, but also validating its viability and effectiveness in a real-world, high-cost scenario through successful pre-training on models up to the 7B parameter scale. For academic research outside of large corporate labs, the 7B scale represents a substantial and representative model size, entirely sufficient to demonstrate the scalability and robustness of our method.
To ensure a fair and controlled comparison, we integrated SPAttention and its variants into the OLMoE framework by replacing only the standard self-attention module, keeping all other aspects of the model architecture and the training pipeline strictly identical. This includes, but is not limited to, the model's parameter count and layer structure, the optimizer state, learning rate schedule, data loading process, and the random seed used for initialization. The same pre-training dataset was used for all models.

\textbf{Training Configuration:} For the OLMoE 0.25B-1.75B series, we use a sequence length of 1024 tokens; for the OLMoE 1B-7B series, we use a sequence length of 4096 tokens. These configurations are sufficient to validate the effectiveness of SPAttention's distance partitioning mechanism and align with standard training practices for the corresponding model scales. We conducted validations at two scales: the OLMoE 0.25B-1.75B series was used for rapid iteration and ablation studies, while the larger OLMoE 1B-7B series served to evaluate our method's effectiveness and scalability.

To ensure a holistic evaluation of model capabilities, we selected a comprehensive and diverse suite of recognized downstream benchmarks. This suite was intentionally designed to probe different aspects of language understanding, including: (1) Commonsense Reasoning (HellaSwag \cite{zellers2019hellaswag}, Winogrande \cite{sakaguchi2020winogrande}, and COPA \cite{roemmele2011choice}) to test implicit world knowledge and reasoning, and (2) Multi-domain Knowledge and Application (MMLU \cite{hendrycks2021measuringmassivemultitasklanguage}) to assess explicit knowledge across STEM, Social Sciences, and Humanities. Furthermore, we evaluate MMLU under both zero-shot and five-shot settings to measure the model's intrinsic knowledge and its in-context learning ability, respectively. This diverse evaluation protocol ensures our findings are robust and generalizable across a variety of linguistic challenges.

We evaluated our full SPAttention method against a Standard Attention baseline and three ablation variants: Local-Only, ExclusiveBands (EBALL), and GappedBands (GBHALF). See the Appendix for training details.

\subsection{Main Results}

We directly compared SPAttention with standard dense attention across two model scales. As presented in Table~\ref{tab:main_results}, on the OLMoE 0.25B-1.75B scale, SPAttention outperforms standard dense attention on four out of six evaluation metrics, with significantly improved overall average performance. This advantage is amplified at the larger OLMoE 1B-7B scale, where SPAttention achieves gains on five out of six metrics with only a minor disadvantage on Winogrande, maintaining a significant lead in overall average performance.

This strongly validates our hypothesis: the over-flexibility of standard MHA leads to redundancy. By introducing a principled structured inductive bias, SPAttention compels functional specialization, liberating resources from redundant local modeling to capture distinct dependencies. This design prevents the model from settling into inefficient local optima, fostering more generalizable representations.

Regarding the few metrics where standard attention holds a slight edge, we attribute this to its ability to ``brute-force'' fit specific distributions with unconstrained capacity, whereas SPAttention's structured design encourages learning fundamental patterns. Holistically, achieving comparable or better performance at roughly half the computational cost unequivocally demonstrates the value of SPAttention as a non-compromise solution.

\subsection{Comparison with Leading Sparse Attention Paradigms}

To provide a comprehensive evaluation against existing sparse attention paradigms, we conducted a systematic comparison against leading sparse attention methods that represent diverse and effective approaches: Longformer \cite{beltagy2020longformerlongdocumenttransformer}, known for its combination of local and global attention; Reformer \cite{kitaev2020reformerefficienttransformer}, which uses locality-sensitive hashing for approximation; and BigBird \cite{NEURIPS2020_c8512d14}, which extends the hybrid approach with random connections. These models are widely recognized as strong and influential baselines in the field of efficient attention. To ensure a rigorously fair comparison and eliminate potential confounding factors from implementation differences or varying levels of kernel optimization, we uniformly implemented all attention mechanisms using FlexAttention \cite{he2024flexattention}, maintaining identical underlying computational frameworks across all methods.

The experimental setup employed an 8-head, 8-layer Transformer with 1024-dimensional hidden states and a maximum sequence length of 1024 tokens. All models were trained for 3000 steps under identical conditions, with performance evaluated on four downstream tasks: HellaSwag \cite{zellers2019hellaswag}, Winogrande \cite{sakaguchi2020winogrande}, COPA \cite{roemmele2011choice}, and STEM, alongside throughput measurement in tokens per second.

The results demonstrate SPAttention's clear superiority across the evaluation landscape. Most notably, SPAttention achieves the highest performance on all four evaluation metrics, with particularly substantial improvements in HellaSwag (+2.4\% over the second-best) and STEM (+0.9\% improvement), culminating in the best overall average performance (0.4048). This comprehensive dominance underscores SPAttention's robust effectiveness across diverse reasoning tasks, from commonsense understanding to domain-specific knowledge application.

\begin{table}[h!]
	\centering
	\footnotesize
	\setlength{\tabcolsep}{2.5pt}
	\begin{tabular}{lcccccc}
		\hline
		\textbf{Method} & \textbf{Hella} & \textbf{Wino} & \textbf{COPA} & \textbf{STEM} & \textbf{Avg} & \textbf{Tok/s} \\ \hline
		Longformer     & 0.2789 & 0.5178 & 0.5700 & 0.2200 & 0.3967 & \textbf{6,340} \\
		Reformer       & 0.2778 & 0.5185 & 0.5600 & 0.2211 & 0.3944 & 6,014 \\
		BigBird        & 0.2772 & 0.5241 & 0.5300 & 0.2250 & 0.3891 & 5,736 \\
		Standard       & 0.2778 & 0.5170 & 0.5700 & 0.2122 & 0.3943 & 3,214 \\
		\textbf{SPAttention} & \textbf{0.2856} & \textbf{0.5263} & \textbf{0.5800} & \textbf{0.2271} & \textbf{0.4048} & 6,167 \\ \hline
	\end{tabular}
	\caption{Performance comparison with sparse attention methods. All throughput tests were conducted in a 4-card NVIDIA A100 environment. Hella=HellaSwag, Wino=Winogrande, Avg=Average, Tok/s=Tokens/second.}
	\label{tab:sparse_comparison}
\end{table}

From an efficiency perspective, SPAttention delivers compelling throughput performance (6,167 tokens/second), ranking second only to Longformer while significantly outpacing the 91.9\% speedup over standard attention. This efficiency-performance combination positions SPAttention as an optimal balance point: it achieves the best task performance while maintaining near-optimal computational efficiency, effectively resolving the traditional trade-off inherent in sparse attention design.

The uniform implementation using FlexAttention \cite{he2024flexattention} eliminates implementation-specific advantages, ensuring that the observed performance gains stem purely from SPAttention's algorithmic innovations rather than engineering optimizations. This controlled comparison validates that SPAttention's Principled Structural Sparsity paradigm fundamentally outperforms existing sparse attention approaches, establishing it as a superior foundation for efficient large language model architectures.

\subsection{Computational Efficiency Analysis}

SPAttention's efficiency stems from a fundamental reduction in computational complexity, which we analyze from both a theoretical and a practical perspective.

\textbf{Theoretical Computational Complexity Breakdown.} Standard Multi-Head Attention is computationally expensive because it performs $H$ independent dense matrix multiplications. The total floating-point operations (FLOPs) for the attention score computation are proportional to $O(H \cdot N^2 \cdot d_k)$, where $d_k$ is the head dimension. In contrast, SPAttention re-architects this process. It transforms the $H$ independent dense computations into a single, partitioned, block-sparse computation. While each head is a distinct computational unit, their collective workload for calculating attention scores sums up precisely to the complexity of a single dense attention operation. The total FLOPs are $\sum_{h=0}^{H-1} O(N \cdot W_h \cdot d_k) = O(N \cdot (\sum W_h) \cdot d_k) = O(N^2 \cdot d_k)$, where $\sum W_h = N$. This constitutes a theoretical reduction in core computational complexity by a factor of $H$, which is the primary source of its acceleration.

\textbf{Overhead and Practical Throughput.} The translation from theoretical FLOPs reduction to practical wall-clock speedup is mediated by several factors. A key source of overhead in any sparse attention mechanism comes from two main areas: 1)Mask Construction and Management, the process of creating and applying the sparsity masks, and 2) Block Computation Pre-processing, which may involve re-shuffling or padding tensors to execute the block-sparse operations efficiently on hardware. Furthermore, according to Amdahl's Law, the total speedup is capped by the proportion of time spent on the attention calculation itself relative to other model components like the Feed-Forward Networks (FFNs).

Despite these factors, SPAttention achieves significant practical speedup. This is because its highly regular and contiguous block-sparse pattern represents an ideal case of algorithm-hardware co-design. Compared to methods with irregular or random sparsity, this regularity minimizes the overhead associated with both mask management and pre-processing. It allows modern hardware and IO-aware kernels to process the computation with maximum efficiency, enabling the theoretical FLOPs reduction to be translated into substantial real-world throughput gains.

\begin{figure*}[t]
	\centering
	\includegraphics[width=0.32\textwidth]{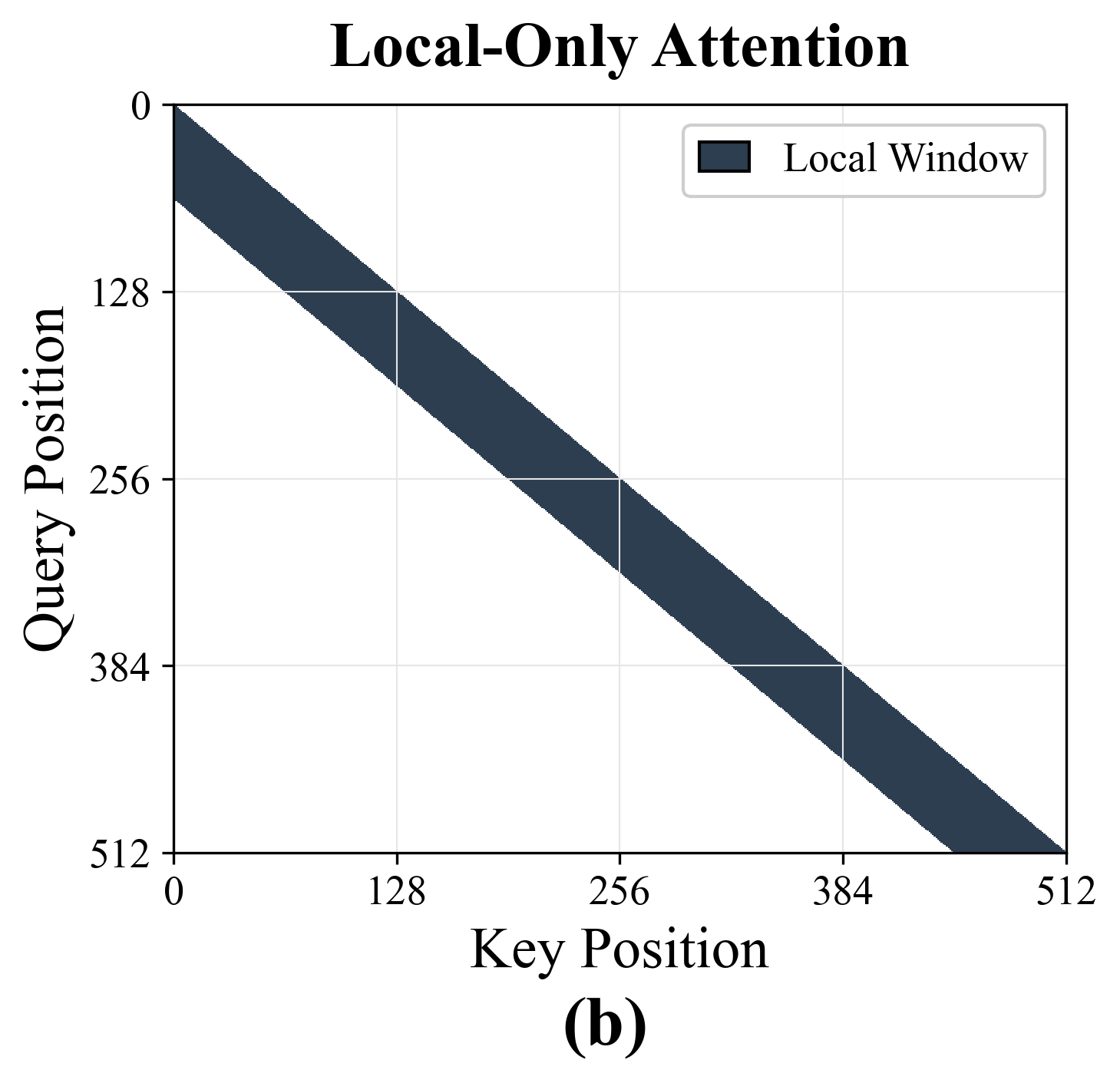} \hfill
	\includegraphics[width=0.32\textwidth]{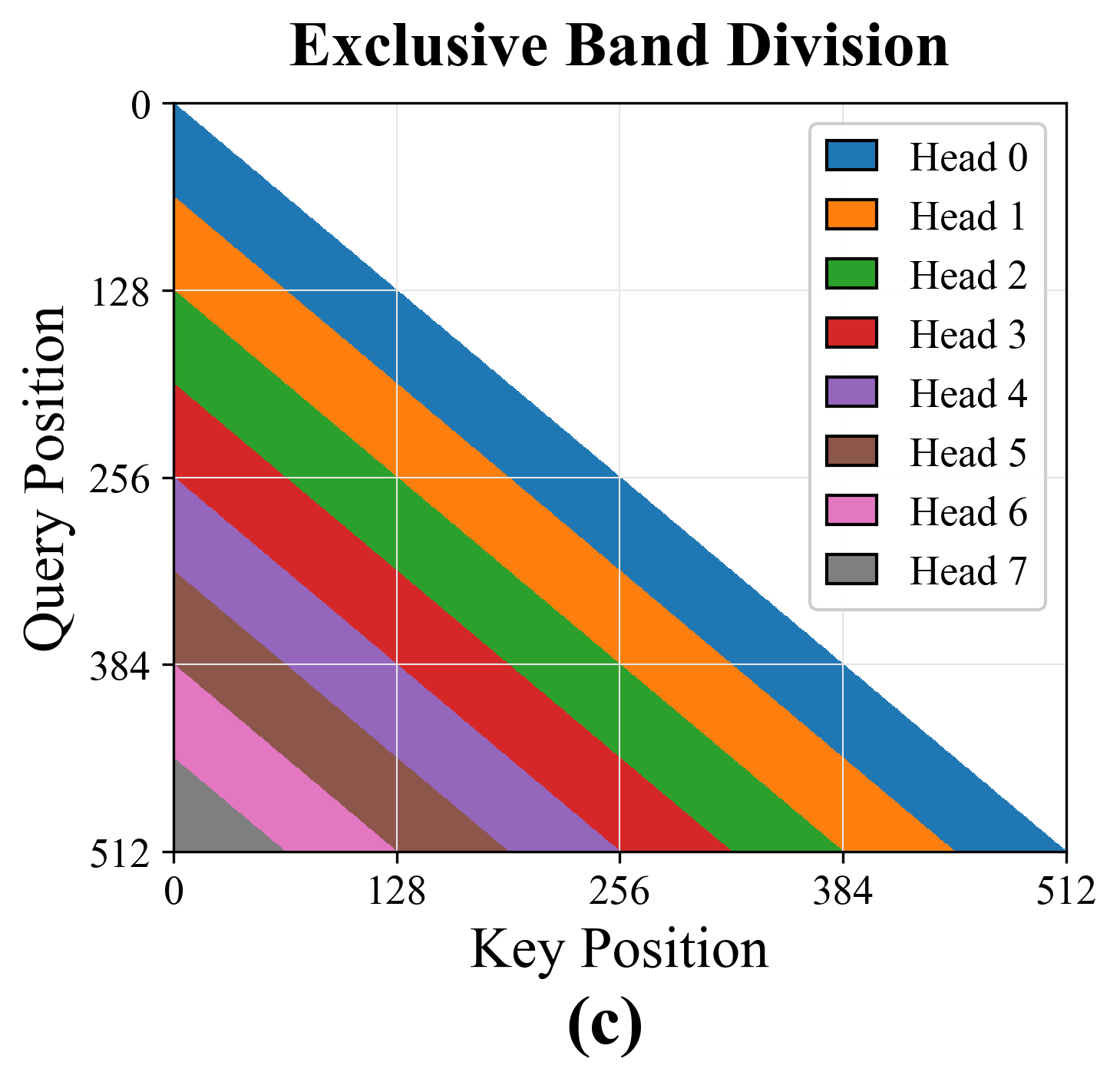} \hfill
	\includegraphics[width=0.32\textwidth]{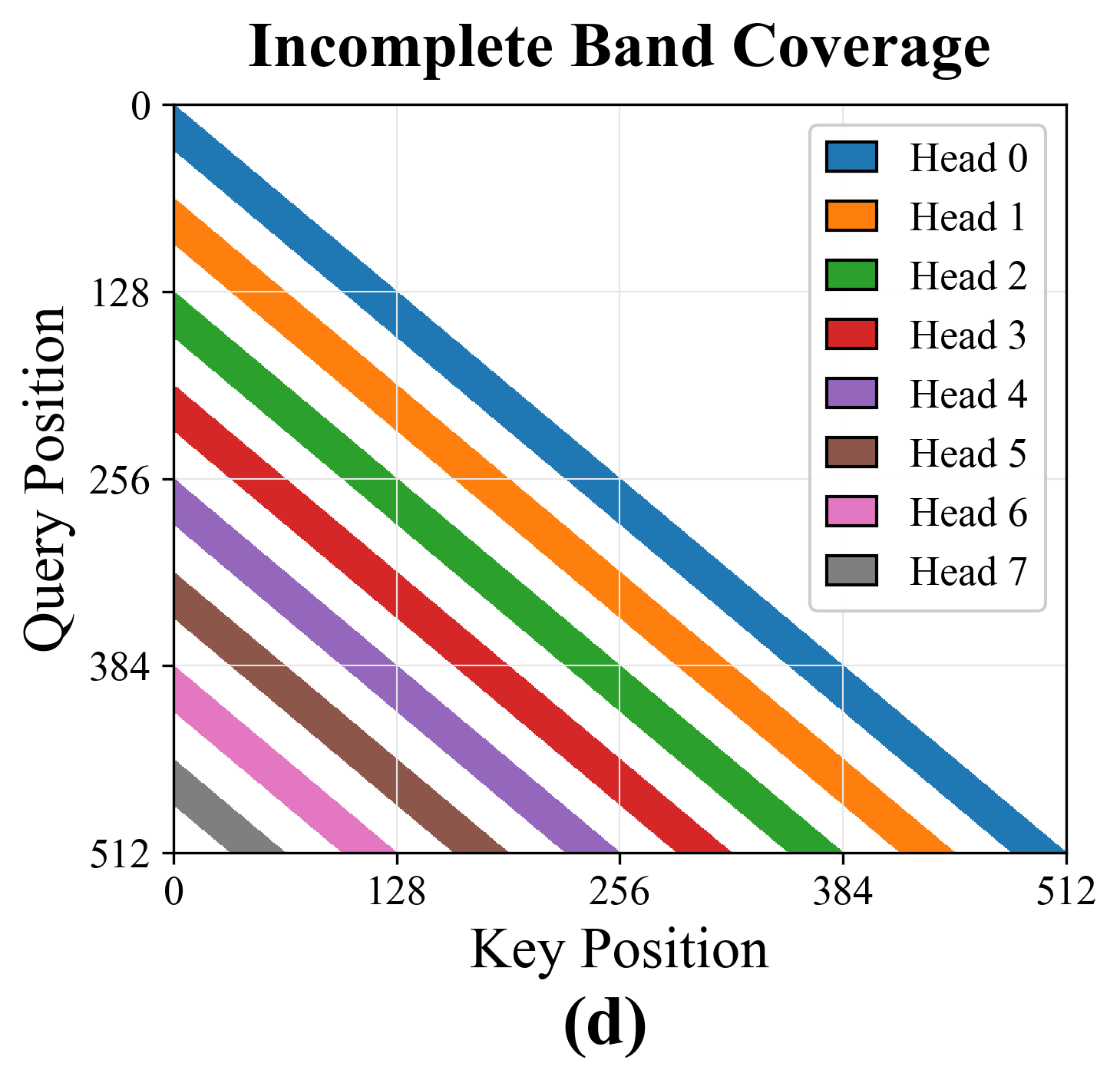}
	\caption{Visualization of the sparse attention patterns for the three ablation variants (H=8, N=1024, showing one representative head for each). From left to right: (a) \textbf{Sliding Window}, representing classical sparse attention where all heads are restricted to a local window with size adapted to sequence length. (b) \textbf{Exclusive Bands (EBALL)}, where local sharing is removed and each head is assigned a unique, non-overlapping distance band. (c) \textbf{Gapped Bands (GBHALF)}, where systematic information blind spots are created by only attending to the first half of each head's assigned region.}
	\label{fig:ablation_patterns}
\end{figure*}

\begin{table*}[t]
	\centering
	\begin{tabular}{lccccccc}
		\hline
		\textbf{Method} & \textbf{Winogrande} & \textbf{COPA} & \textbf{STEM} & \textbf{SocSci} & \textbf{STEM (5-shot)} & \textbf{Humanities (5-shot)} & \textbf{Average} \\ \hline
		GBHALF       & 0.5099 & 0.5200 & 0.2599 & 0.2306 & 0.2349 & 0.2336 & 0.3315 \\
		EBALL        & 0.5091 & 0.4800 & 0.2577 & \textbf{0.2361} & 0.2548 & 0.2353 & 0.3289 \\
		Sliding Window   & 0.5178 & 0.5300 & \textbf{0.2688} & 0.2128 & 0.2333 & 0.2425 & 0.3376 \\ \hline
		SPAttention  & \textbf{0.5272} & \textbf{0.5400} & 0.2661 & 0.2221 & \textbf{0.2608} & \textbf{0.2440} & \textbf{0.3434} \\ \hline
	\end{tabular}
	\caption{Ablation Study Results on the OLMoE 0.25B-1.75B scale. Best results for each metric are highlighted in bold.}
	\label{tab:ablation}
\end{table*}

To validate this, we measured the wall-clock time for a complete forward-backward pass using a batch size of 1, 8 attention heads, a sequence length of 4096, and a head dimension of 128. Our empirical results confirm the theoretical advantages, showing that SPAttention achieves an approximately 2$\times$ training speedup compared to a standard dense attention baseline under identical conditions.

\subsection{Ablation Studies}

To investigate the contribution of each component within SPAttention's design and demonstrate its advantages over classical sparse attention methods, we conducted exhaustive ablation studies on the OLMoE 0.25B-1.75B scale.

\textbf{Rationale for Ablation Design}: To systematically validate the contribution of each component in SPAttention's design, we designed three specific ablation variants, including a comparison with classical sparse attention methods. We chose the most representative sliding window sparse attention as a baseline comparison, as it is the most classical and widely adopted method in the sparse attention field. Through carefully designed ablation variants, we can directly quantify the roles of core design elements such as balanced distance partitioning and complete information pathways, while demonstrating our method's advantages over classical sparse attention.

The comparison involved four methods: (1) a Standard Attention baseline; (2) complete SPAttention method; and (3) three ablation variants: Sliding Window, ExclusiveBands (EBALL), and GappedBands (GBHALF). The attention patterns of our ablation variants are visualized in Figure~\ref{fig:ablation_patterns}.

The ablation study results, summarized in Table \ref{tab:ablation}, unequivocally demonstrate that every component of SPAttention's full design is crucial for its success. As shown in the table, the complete SPAttention achieves the highest overall average score and outperforms all variants and baselines on multiple key individual metrics.

The degraded performance of the variants validates SPAttention's core design principles. The classical Sliding Window approach struggles on tasks requiring broad context, lacking a long-range mechanism. The GappedBands (GBHALF) model fails due to information blind spots, underscoring the need for a complete information pathway, while the unbalanced allocation in ExclusiveBands (EBALL) impairs dependency capture, proving the value of our balanced partitioning strategy.

These results demonstrate that SPAttention's superior performance stems not from a single trick, but from the synergistic outcome of its core principles: completeness, exclusivity, and balance. Any deviation from this holistic design incurs a significant performance penalty, establishing SPAttention as a robust general-purpose architecture.

\section{Conclusion}

This paper has confronted the core conflict within the self-attention mechanism: the tension between its powerful expressivity and its $O(H \cdot N^2)$ computational complexity, which is fraught with structural redundancy. To resolve the long-standing efficiency-performance trade-off, we introduced SPAttention, a novel attention paradigm based on Principled Structural Sparsity. The core philosophy of SPAttention is not to merely drop connections, but to actively reorganize and specialize the computational workload by partitioning the attention distance spectrum into balanced, non-overlapping bands. This design transforms the attention mechanism from $H$ redundant, independent computations into a single, collaborative $O(N^2)$ computation.

Our empirical validation demonstrates that SPAttention performs on par with or better than standard dense attention and consistently outperforms leading sparse methods. Our work's crucial insight is that thoughtfully designed structural sparsity serves as a powerful inductive bias, simultaneously improving computational efficiency and model capabilities by eradicating redundancy at its source. This paves a promising path for pruning~\cite{hu2025mosaicpruninghierarchicalframework} and designing next-generation models like LLaMA~\cite{touvron2023llama} and Mamba~\cite{gu2023mamba} while mitigating their expensive pre-training costs.

\section{Acknowledgments}
This work was supported by the National Natural Science Foundation of China (Grant Nos. 62572389, 72293581, 72274152, 62402376).

\bibliography{aaai2026}

\end{document}